\documentclass{article}
\usepackage[preprint,nonatbib]{neurips_2023}

\usepackage[utf8]{inputenc} % allow utf-8 input
\usepackage[T1]{fontenc}    % use 8-bit T1 fonts
\usepackage{hyperref}       % hyperlinks
\usepackage{url}            % simple URL typesetting
\usepackage{booktabs}       % professional-quality tables
\usepackage{multirow}
\usepackage{amsfonts}       % blackboard math symbols
\usepackage{nicefrac}       % compact symbols for 1/2, etc.
\usepackage{microtype}      % microtypography
\usepackage{lipsum}
\usepackage{bbm}
\usepackage{fancyhdr}       % header
\usepackage{graphicx}       % graphics
\usepackage{subcaption}     % subcaption

\usepackage{wrapfig}
\usepackage{hyperref}
\usepackage{color}
\usepackage{url}
\usepackage{enumitem}
\usepackage{graphicx}
\usepackage{multirow}
\usepackage{booktabs}
\usepackage{adjustbox}  % for adjustments
\usepackage{dblfloatfix}
\usepackage{algorithm}% http://ctan.org/pkg/algorithm
\usepackage{algpseudocode}% http://ctan.org/pkg/algorithmicx
\usepackage{amssymb}
\usepackage{minitoc}

\usepackage{makecell}
\renewcommand{\paragraph}[1]{ \noindent \textbf{#1}}

\title{Sparse Fine-tuning for Inference Acceleration \\ of Large Language Models
}

\author{
  Eldar Kurtic\thanks{These authors contributed equally.} \\
  IST Austria \\
  \texttt{eldar.kurtic@ista.ac.at} \\
  \And
  Denis Kuznedelev\footnotemark[1] \\
  Skoltech \& Yandex \\
  \texttt{denis.kuznedelev@skoltech.ru} \\
  \And
  Elias Frantar \\
  IST Austria \\
  \texttt{elias.frantar@ista.ac.at} \\
  \And
  Michael Goin \\
  Neural Magic \\
  \texttt{michael@neuralmagic.com} \\
  \And
  Dan Alistarh \\
  IST Austria \& Neural Magic \\
  \texttt{dan.alistarh@ista.ac.at} \\
}

\begin{document}
\maketitle

\begin{abstract}
    We consider the problem of accurate \emph{sparse fine-tuning} of large language models (LLMs), that is, fine-tuning pretrained LLMs on specialized tasks, while inducing sparsity in their weights. 
    On the accuracy side, we observe that standard loss-based fine-tuning may fail to recover accuracy, especially at high sparsities. To address this, we perform a detailed study of distillation-type losses, determining an L2-based distillation approach we term SquareHead which enables accurate recovery even at higher sparsities, across all model types. 
    On the practical efficiency side, we show that sparse LLMs can be executed with speedups by taking advantage of sparsity, for both CPU and GPU runtimes. 
    While the standard approach is to leverage sparsity for computational reduction, we observe that in the case of memory-bound LLMs sparsity can also be leveraged for reducing memory bandwidth. We exhibit end-to-end results showing speedups due to sparsity, while recovering accuracy, on T5 (language translation), Whisper (speech translation), and open GPT-type (MPT for text generation).      
    For MPT text generation, we show for the first time that sparse fine-tuning can reach 75\% sparsity without accuracy drops, provide notable end-to-end speedups for both CPU and GPU inference, and highlight that sparsity is also compatible with quantization approaches. Models and software for reproducing our results are provided in Section~\ref{sec:reproducibility}. 
\end{abstract}

\vspace{-1em}
\section{Introduction}
\vspace{-0.3em}

Large Transformer models~\cite{vaswani2017attention} have gained high popularity and adoption due to their breakthrough performance across a wide range of challenging tasks. 
To address their high runtime costs, several acceleration techniques have been developed, e.g.,~\cite{dao2022flashattention, frantar2022gptq, dettmers2022optimizers, dettmers2022case}. 
For inference acceleration, the most popular technique is \emph{quantization}, e.g.,~\cite{dettmers2022llm,xiao2022smoothquant,frantar2022gptq,dettmers2022case}:  
 LLMs can be quantized down to 4 bits per weight (or more) almost without loss, and that this can be leveraged for  inference speedups, e.g~\cite{frantar2022gptq, dettmers2022case}. 
Yet, quantization methods are reaching accuracy limits at around 3 bits per weight: at this point, accuracy appears hard to recover with current techniques~\cite{yao2022zeroquant, dettmers2023spqr, chee2023quip}. 

A key compression alternative to quantization is \emph{weight sparsity}~\cite{lecun1990optimal}, which consists of pruning individual LLM connections by setting them to zero. For smaller models, e.g., BERT~\cite{devlin2018bert}, it is known~\cite{2020-sanh, kurtic2022optimal} that high levels of sparsity can be applied during \emph{fine-tuning}, i.e. the process by which the pretrained model is adapted to a ``downstream'' task, such as question answering or text classification, leading to significant speedups. However, it is not known whether similar techniques work at the scale of LLMs. 

In this preliminary work, we study efficient sparse fine-tuning for LLMs, across three modern applications: speech transcription using Whisper~\cite{radford2023robust}, specialized to a specific language, machine translation using T5~\cite{wei2021finetuned}, specialized to a specific language pair~\cite{machavcek2014results}, and higher-level reasoning using the open GPT-type MPT model~\cite{mpt}, specialized on the Grade-School Math (GSM) task~\cite{cobbe2021training}.  

In this context, our contributions are as follows: 

\begin{itemize}[leftmargin=0.5cm]
    \item We observe that naive sparse fine-tuning~\cite{2020-sanh}, which follows dense fine-tuning while gradually imposing sparsity, is challenging to apply for LLMs due to training instability, which manifests itself in several forms: 
  1) \emph{loss spikes} at higher sparsity values, leading to divergence, 2) \emph{poor recovery}, as the relatively small amount of fine-tuning data available for the subtask may not be sufficient to recover accuracy; or 3) \emph{overfitting}, since iterating multiple times over the limited fine-tuning data leads to low training loss, but high validation error. 
    \item To address this, we investigate fine-tuning sparse models obtained via SparseGPT~\cite{frantar2023massive} using various losses which incorporate standard cross-entropy, output knowledge distillation~\cite{hinton2015distilling}, and a type of per-token $\ell_2$ knowledge distillation inspired by~\cite{sun2019patient, kurtic2023ziplm, frantar2022spdy} which we call \emph{SquareHead}. We show that SquareHead distillation consistently recovers accuracy, even at high sparsities. 
    \item On the practical side, we show that the resulting sparse models can be executed with inference speedups. For CPU inference, we leverage   the DeepSparse inference runtime~\cite{deepsparse} to obtain speedups across all three applications. In addition, focusing on generative inference, we present GPU speedups via a GPU-aware N:M sparse format, and remarkable CPU speedups, e.g. of 7.5x with minimal accuracy loss vs the FP32 baseline, jointly leveraging sparsity and quantization. 
    
\end{itemize}

Together, our results show that, with the right set of techniques, sparsity can be successfully applied in the challenging LLM fine-tuning scenario, and that this can lead to significant speedups across practical scenarios, both on CPUs and GPUs.

\vspace{-0.5em}
\section{Methodology}
\label{sec:method}

\vspace{-0.5em}
\subsection{Sparse Fine-tuning}

\paragraph{Sparsification.}
To obtain a set of sparse models satisfying target compression requirements, we increase the sparsity level gradually while fine-tuning the model on the task of interest. Unless otherwise stated, we start from a list of desired sparsity levels, in increasing order, and iteratively prune and fine-tune the model while following the original fine-tuning recipe in each subsequent cycle.

\paragraph{Distillation strategies.}
LLMs are notoriously difficult to train and fine-tune, and we found this to be particularly the case when sparsity is imposed during an often short fine-tuning cycle. 
Choosing the ``right'' loss function for fine-tuning is particularly important, and thus we investigate knowledge distillation (KD) approaches. 
The sparse student model is trained to mimic the 
behavior of a dense teacher, which has already been fine-tuned on the target task. The most common KD strategy adds a loss term measuring the KL divergence between student and teacher outputs~\cite{hinton2015distilling}. 
However, we obtain better results by going further and distilling intermediate representations. 

\emph{Standard output distillation} uses the KL-divergence between student and teacher logits as loss:
\begin{equation}
\mathcal{L}_{\mathrm{logit}} = D_{\mathrm{KL}} (\theta_t || \theta_s) = 
\frac{1}{\sum \limits_{i}^{B \times seq} \mathbbm{1}  \left[i \notin \mathbf{P} \right]}
\sum \limits_{i}^{B \times seq} \mathbbm{1} \left[i \notin \mathbf{P} \right] p_{\theta_t} (\mathbf{x}_i) \log \frac{p_{\theta_t} (\mathbf{x}_i)}{p_{\theta_s} (\mathbf{x}_i)}.
\end{equation}

Above, $B$ stands for batch size, $seq$ for sequence length, $p_{\theta_t} (\mathbf{x}_i)$ and 
$p_{\theta_s} (\mathbf{x}_i)$ denote output probabilities for teacher and student model respectively. 
The notation $\mathbbm{1}  \left[i \notin \mathbf{P} \right]$ means that the loss for padding tokens $\mathbf{P}$ is discarded. 

\emph{To transfer intermediate representations}, we examine normalized mean squared error (MSE) loss on each feature representation, which we for simplicity call \textbf{SquareHead}: 

    \begin{equation}
    \mathcal{L}_{\mathrm{feat}}^{l} =
    \frac{\textrm{MSE}(f_{t}^{l}, f_{s}^{l})}
    {\textrm{MSE}(f_{t}^{l}, 0)}, 
    \quad \mathrm{shape}(f_{s}^{l}) = \mathrm{shape}(f_{t}^{l}) = B \times seq \times d_{\mathrm{model}}.
    \end{equation}

Here, $f_{t}^{l}$ and $f_{s}^{l}$ denote the feature map of $l$-th layer of teacher and student model, and $\textrm{MSE}$ represents the mean squared error calculated as $\textrm{MSE}(X, Y) = \frac{1}{N} \sum_{i=0}^{N} (x_i - y_i)^2$, for $N$-dimensional vectors $X$ and $Y$. We discard terms that correspond to padding tokens. The motivation 
for the normalization is that the magnitude of activations may greatly vary between different layers, thus focusing the 
optimization procedure to optimize the layers with largest norm. We observed that for some models MSE loss without normalization leads to 
instabilities in training. The total feature loss is the sum of per-layer losses over all encoder/decoder blocks.

\paragraph{SquareHead Distillation.} 
The overall SquareHead distillation loss is the sum of original task loss and SquareHead term with equal weights: $\mathcal{L} = \mathcal{L}_{\mathrm{task}}  
+ \mathcal{L}_{\mathrm{feat}}.$

\paragraph{Prior Work.} As stated previously, variants of this loss have been used by~\cite{sun2019patient, kurtic2023ziplm, frantar2022spdy} in the context of compressing smaller-scale models during the fine-tuning process, such as BERT-style models~\cite{devlin2018bert} for question-answering or sentiment classification fine-tuning tasks. 
In the context of BERT fine-tuning, it is known that variants of knowledge distillation can help reduce accuracy loss during fine-tuning, e.g.~\cite{2020-sanh, kurtic2022optimal, frantar2022spdy, kurtic2023ziplm, xia2022structured}. Relative to this line of work, in this paper we determine a type of distillation loss that is consistently effective for accurate sparse fine-tuning of large models.

\vspace{-0.5em}
\subsection{Runtime Acceleration from Sparsity}

We now turn our attention to the question of obtaining practical runtime speedups from a sparse fine-tuned model. 
The standard strategy is to leverage sparsity \emph{reducing computational load}: since we can skip multiplications with zero, higher sparsity means lower inference computation. However, leveraging the relatively-low sparsities arising in accurate deep models for computational gains is known to be difficult, especially on GPUs~\cite{gale2020sparse, hoefler2021sparsity}.     

\begin{wrapfigure}{R}{0.4\textwidth}
\centering
\includegraphics[width=0.4\textwidth]{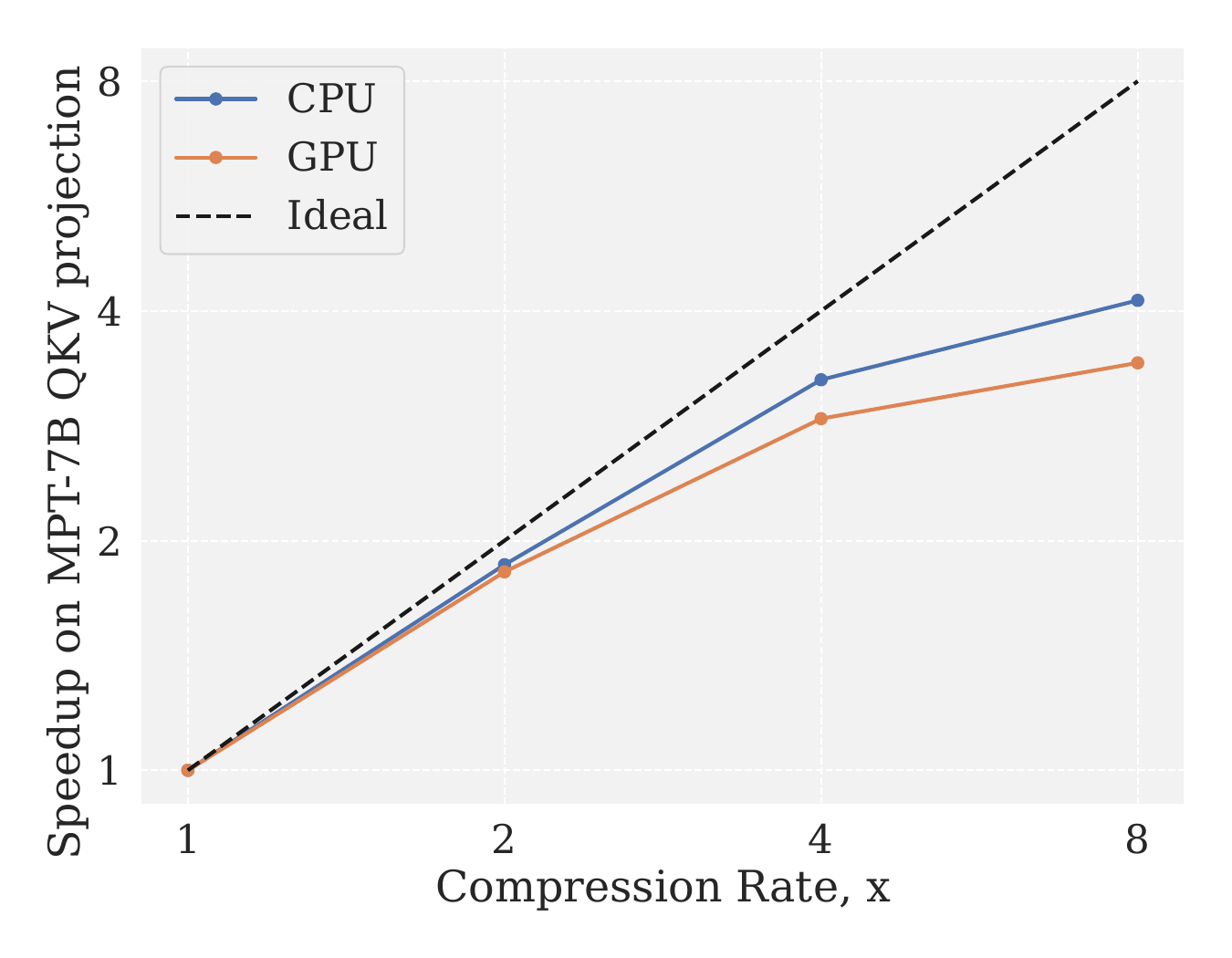}
\caption{\label{fig:self-speedups} Speedups for sparse CPU and GPU kernels on an MPT-7B layer.} \vspace{-1.5em}
\end{wrapfigure}
In the case of LLMs, we have a complementary path to sparse acceleration, since these models are often \emph{memory bound}~\cite{frantar2022gptq, dettmers2022case}. That is, a significant fraction of the generation runtime can be taken by loading the weights (from memory to registers) to perform layer computation. 

To leverage this, we can store sparse weights in compressed form, and decompress them on-the-fly while performing the layer computation. This may even involve multiplication with zeros, as long as they do not have to materialize weights in slow memory. 
We present results on CPU and GPU runtimes which leverage variants of this approach. For CPU execution, we leverage the DeepSparse runtime~\cite{deepsparse} which implements sparsity-aware inference in both memory-bound and compute-bound scenarios.

\paragraph{GPU Implementation Sketch.} For GPU inference, we show speedup potential via the following implementation for ``uniform'' sparsity patterns such as 16:32 (50\%), 16:64 (75\%) and 16:128 (87.5\%). (For clarity, 16:32 means 16 non-zeros in every block of 32 consecutive weights.) 
We implement a custom CUDA kernel where each threadblock fetches tiles of 16 $\times$ \#threads weights to shared memory and then accumulates partial matrix-vector product outputs by unpacking the corresponding INT32 sparsity bitmasks to determine when weights should be multiplied with inputs. 

Executed on an NVIDIA A6000 GPU for a $4096\times12288$ matrix (shapes of the QKV projection layer in MPT-7B model), our kernel yields $1.82\times$ speedup over dense FP16 execution, which is close to the theoretical expectation of $1.77\times$ (equivalent to 9 bits per weight) for a bitmask representation of sparsity. 

We show self-speedup results, i.e. versus the respective dense baseline, for this $4096\times12288$ matrix  both on CPU and GPU in Figure~\ref{fig:self-speedups}, observing that both implementations achieve substantial speedups. For reference, we can prune the generative MPT-7B model both in 16:32 and 16:64 formats, and fully recover accuracy on GSM8K (see Table ~\ref{tab:NM_mpt} in Section~\ref{sec:mpt}).

\vspace{-0.5em}
\section{Application 1: Compression of T5 and Whisper Translation Models}
\vspace{-0.5em}

\paragraph{Experimental Setup.} 
We consider sparsity levels corresponding to 2x, 3x, 4x, 5x, 6x, 7x, 8x, and 10x compression ratios, applied uniformly per layer with the SparseGPT~\cite{frantar2023massive} pruner. We compare sparse fine-tuning with three variants of losses.
Cross Entropy (original loss, $\mathcal{L}_{\mathrm{task}}$), Standard KD ($\mathcal{L}_{\mathrm{task}} + \mathcal{L}_{\mathrm{logit}}$), and SquareHead KD ($\mathcal{L}_{\mathrm{task}} + \mathcal{L}_{\mathrm{feat}}$). We report  hyperparameters in Appendix~\ref{appendix:training_hyperparameters}.
Speedups are reported for end-to-end execution in the DeepSparse inference engine~\cite{deepsparse} on an Intel Sapphire Rapids CPU with 8 cores (AWS m7i.4xlarge). In the following section, the comparison between distillation strategies (left figure) is done \emph{without} pruning the final output head, whereas 
the language model head is pruned in the runtime experiments (right tables), since this has non-neglible impact on total inference time. 

\begin{figure}[!h]
    \centering
    
    % First minipage for the figure
    \begin{minipage}[b]{0.45\linewidth}
        \centering
        \includegraphics[width=0.9\linewidth]{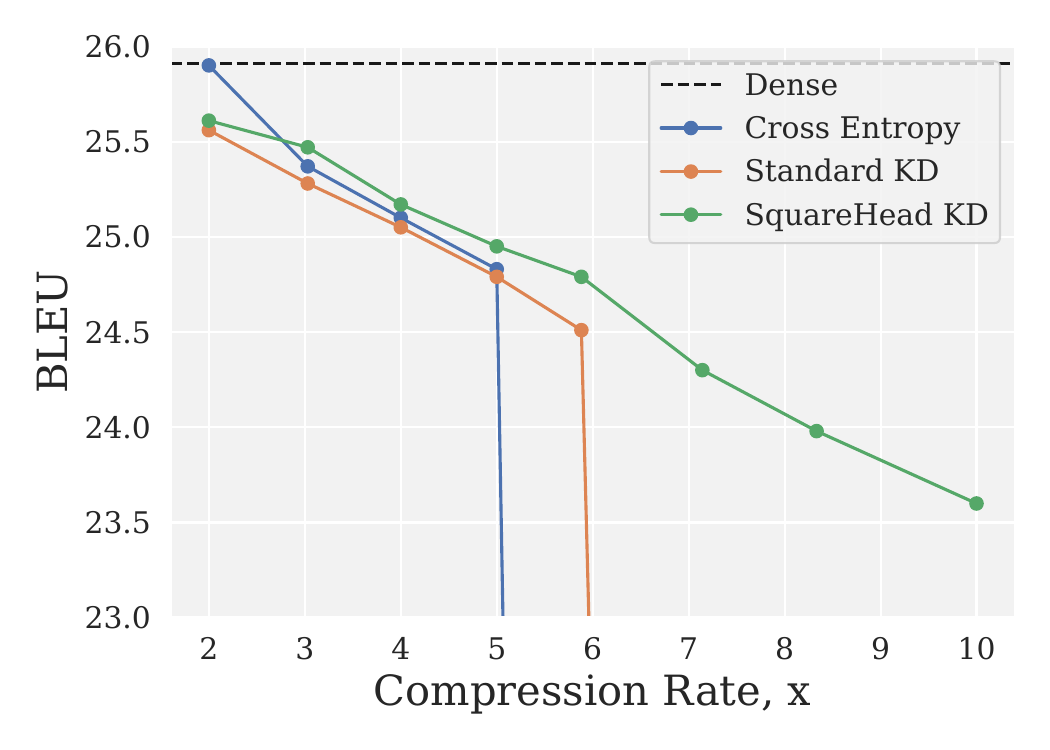}
        \captionof{figure}{
            BLEU ($\uparrow$) score for variants of loss functions on English-German WMT14 and T5-Small at various compression rates. 
        }
        \label{fig:t5-comparison}
    \end{minipage}
    \hfill  % Creates a horizontal space between the minipages
    % Second minipage for the table
    \begin{minipage}[b]{0.45\linewidth}
        \centering
        \small{
        \setlength\tabcolsep{1.7pt}
        \begin{tabular}{cccc}
            \toprule
            Model & Sparsity (\%) & BLEU ($\uparrow$) & Speedup $(\times)$ \\
            \midrule
            \multirow{10}{*}{T5} & 0 & 25.91  & 1.00 \\
            & 50 & 25.44 & 1.89 \\
            & 67 & 25.04 & 2.12 \\
            & 75 & 24.67 & 2.14 \\
            & 80 & 24.57 & 2.16 \\
            & 83 & 24.31 & 2.22 \\
            & 86 & 24.11 & 2.25 \\
            & 88 & 23.47 & 2.38 \\
            & 90 & 23.10 & 2.42 \\
            \bottomrule
        \end{tabular}
        }
        \captionof{table}{BLEU score and speedups of T5-Small for various sparsities and SquareHead KD loss on English-German WMT14.
        } 
        \label{tab:compression_t5}
    \end{minipage}

\end{figure}

\paragraph{Compression for language translation using T5.} 
We begin by investigating sparsification of a pretrained T5 model \cite{raffel2020exploring}, fine-tuned on the popular English-German subset of WMT14~\cite{bojar-etal-2014-findings}. Following standard practice, 
we compute BLEU scores on a validation set of the corresponding split as a measure of model's accuracy. Figure~\ref{fig:t5-comparison} shows accuracy-vs-sparsity trade-off for various loss functions. Table~\ref{tab:compression_t5} shows scores for the best-performing loss variant (SquareHead KD), together with self-speedups. The dense baseline needs 370ms to encode/decode 128 tokens. 

We observed that sparse training with Cross Entropy only and Standard KD becomes unstable at high sparsity and only SquareHead KD manages to produce highly-sparse models with reasonable performance. An example of training loss curve is presented in Appendix~\ref{appendix:t5_divergence}.

\begin{figure}[!h]
    \centering
    
    % First minipage for the figure
    \begin{minipage}[b]{0.45\linewidth}
        \centering
        \includegraphics[width=0.9\linewidth]{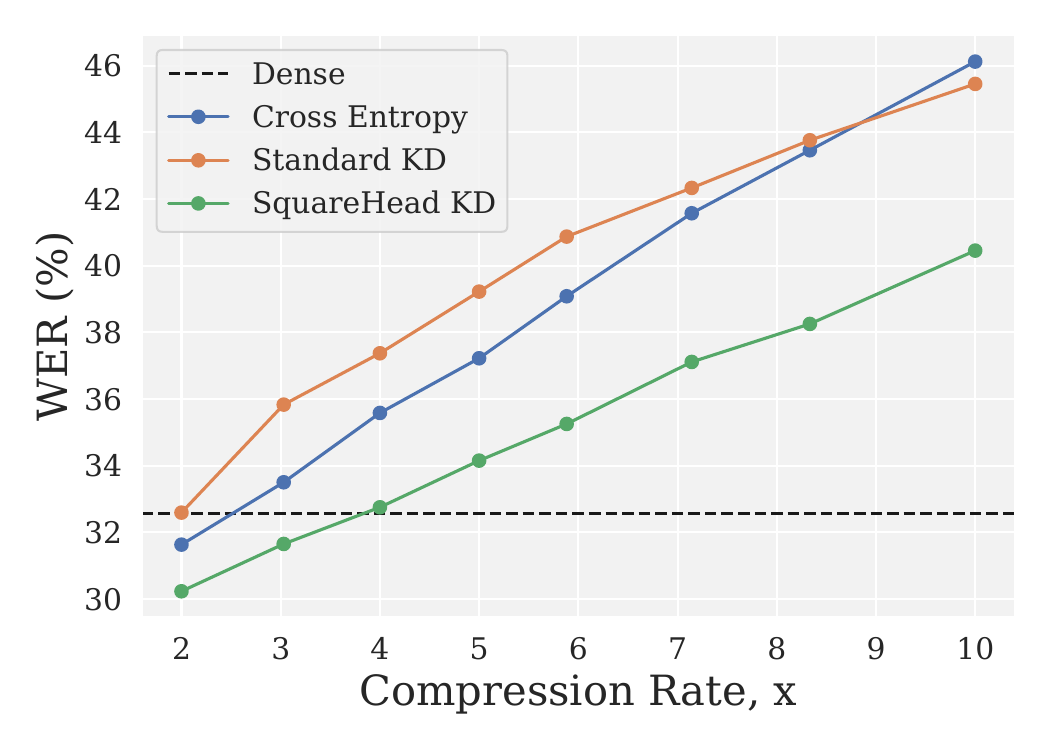}
        \captionof{figure}{
            WER ($\downarrow$) for variants of loss functions on Hindi dataset and Whisper-Small at various compression rates.
        }
        \label{fig:one_shot_pruning_ng}
    \end{minipage}
    \hfill 
    % Creates a horizontal space between the minipages
    % Second minipage for the table
    \begin{minipage}[b]{0.45\linewidth}
        \centering
        \small{
        \setlength\tabcolsep{1.7pt}
        \begin{tabular}{cccc}
            \toprule
            Model & Sparsity (\%) & WER ($\downarrow$) & Speedup $(\times)$ \\
            \midrule
            \multirow{10}{*}{Whisper} & 0 & 32.57 & 1.00  \\
            & 50 & 30.87 & 1.58 \\
            & 67 & 31.85 & 2.15 \\
            & 75 & 33.14 & 2.23 \\
            & 80 & 34.69 & 2.35 \\
            & 83 & 35.77 & 2.49 \\
            & 86 & 37.33 & 2.52 \\
            & 88 & 38.64 & 2.58 \\
            & 90 & 40.60 & 2.67 \\
            \bottomrule
        \end{tabular}
        }
        \captionof{table}{WER and speedups of Whisper-Small for various sparsities and SquareHead KD loss on Hindi dataset.} 
        \label{tab:compression_whisper}
    \end{minipage}

\end{figure}

\paragraph{Compression of speech-to-text using Whisper.}
In Figure~\ref{fig:one_shot_pruning_ng} and Table~\ref{tab:compression_whisper}, we study compression of Whisper-Small (244M)~\cite{radford2022robust} for Automatic Speech Recognition (ASR) on the Hindi (Hi) language subset of CommonVoice 11.0~\cite{ardila-etal-2020-common}. We report Word Error Rate (WER) as a standard measure of ASR performance. 
Our dense CPU baseline takes 882ms to transcribe an audio sequence of 15 seconds and a detailed breakdown is presented in Appendix~\ref{appendix:t5_whisper_perf}. 

\paragraph{Discussion.} 
Results show that
the SquareHead KD loss improves upon standard approaches in these two fine-tuning scenarios, especially at higher sparsities, where the latter tend to diverge. 
(We provide an analysis of this phenomenon in Appendix~\ref{appendix:entropy}.) 
Overall, we can induce moderate sparsity levels (50-70\%) while preserving or even improving accuracy, leading to speedups of 1.5-2x. We can obtain higher speedups, of up to 2.67x, with 80-90\% sparsity, at the price of higher error rates. 

\vspace{-0.5em}
\section{Application 2: Compression of Generative Models}\label{sec:mpt}
\vspace{-0.5em}

Next, we investigate compression of generative models, specifically the open-source GPT-style Mosaic pretrained model MPT-7B~\cite{mpt}. We focus on GSM8K~\cite{gsm8k}, a dataset with high quality and diverse grade school math problems, following the recipe provided by~\cite{Anyscale}. On this task, in zero-shot mode, the baseline model completely fails, with a score of 0\%, whereas in 8-shot evaluation it scores only 6.8\%. These results suggest that the model necessitates additional refinement via supervised fine-tuning (SFT).

\paragraph{Experimental setup.} First, we fine-tune MPT-7B via SFT to obtain a highly accurate and competitive dense baseline, which we use as the teacher in knowledge distillation runs. Then, we apply oneshot unstructured pruning with SparseGPT to 40\%, 50\%, 60\%, 70\%, and 80\% sparsity targets, uniformly across all layers, which correspond to 1.7x, 2.0x, 2.5x, 3.3x, and 5.0x compression ratios respectively. We explore how different sparse fine-tuning techniques help in recovering accuracy of the dense baseline model. We provide detailed description of hyper-parameters in Appendix~\ref{appendix:mpt_hyperparameters}. After fine-tuning, we investigate compatibility of sparse models with quantization via post-training quantization to INT8. To achieve this we leverage the SparseML~\cite{pmlr-v119-kurtz20a} library and quantize to 8-bits the weights and activations of all linear weight matrices, and two batch matrix multiplications in attention layers.
For accuracy evaluation on the GSM8K task we utilize the standardized evaluation protocol via Language Model Evaluation Harness~\cite{lmeval}.\looseness=-1

\begin{figure}[!t]
    \centering
    
    % First minipage for the figure
    \begin{minipage}{0.48\linewidth}
        \centering
        \includegraphics[width=0.9\linewidth]{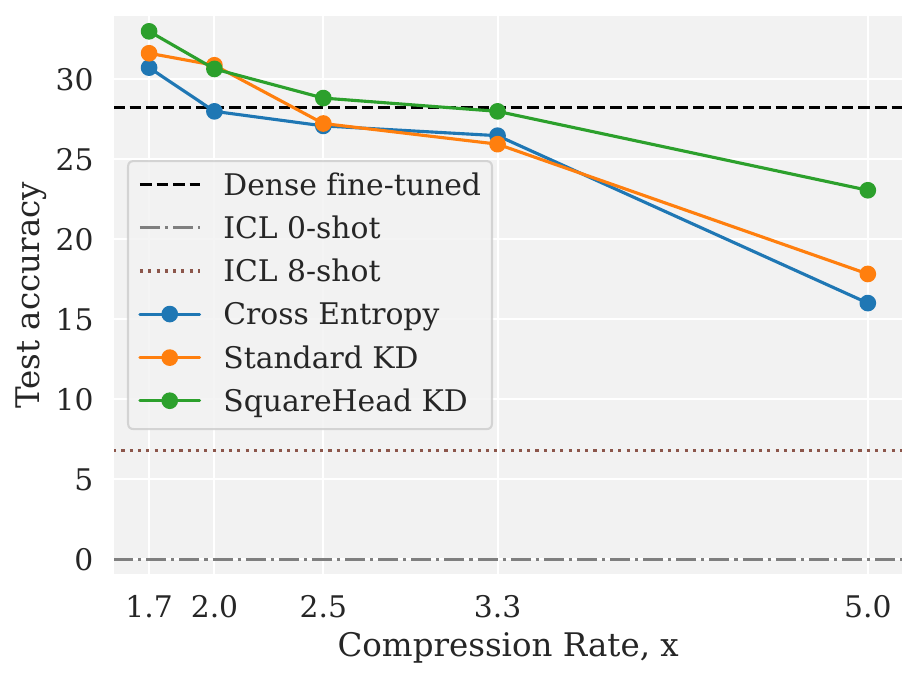}
        \captionof{figure}{
            Test accuracy for variants of loss functions on GSM8K dataset and MPT-7B model at various compression rates. ICL stands for in-context learning results using the pretrained dense model, without fine-tuning.
        }
        \label{fig:mpt-comparison}
    \end{minipage}
    \hfill  % Creates a horizontal space between the minipages
    % Second minipage for the table
    \begin{minipage}{0.48\textwidth}
    \centering
        \small{
        \setlength\tabcolsep{2pt}
            \begin{tabular}{c|cc|cc}
                \toprule
                 & \multicolumn{2}{|c|}{FP32} & \multicolumn{2}{c}{INT8} \\
                \midrule
                \makecell{Sparsity\\(\%)} & \makecell{Test\\accuracy} &  \makecell{CPU\\speedup} & \makecell{Test\\accuracy} &  \makecell{CPU\\speedup} \\
                \midrule
                 0  & $28.2$ &  1.00x & $27.8$ & 3.98x \\
                 40 & $32.9$ &  1.54x & $30.3$ & 5.31x \\
                 50 & $30.6$ &  1.78x & $30.7$ & 5.72x \\
                 60 & $28.8$ &  2.07x & $28.4$ & 6.70x \\
                 70 & $28.0$ &  2.62x & $27.1$ & 7.49x \\
                 80 & $23.1$ &  3.35x & $21.1$ & 9.08x \\
                \bottomrule
            \end{tabular}
        \captionof{table}{Test accuracy of pruned (FP32) and pruned-quantized (INT8) MPT-7B models on the GSM8K dataset. Speedups are measured relative to FP32, for end-to-end decode latency in  DeepSparse~\cite{deepsparse} using an 8 core CPU (AWS m7i.4xlarge) at sequence length 512.}
        \label{tab:compression_mpt}
        }
    \end{minipage}
\end{figure}

\begin{minipage}{0.35\textwidth}
\centering
    \small{
    \setlength\tabcolsep{2pt}
    \begin{tabular}{c|c|c}
        \toprule
        \makecell{N:M\\format} & \makecell{Sparsity\\(\%)} & \makecell{Test\\accuracy} \\
        \midrule
         dense & 0  & $28.2$ \\
         2:4 & 50 & $31.4$ \\
         16:32 & 50 & $31.4$ \\
         16:64 & 75 & $28.6$ \\
        \bottomrule
    \end{tabular}
    \captionof{table}{Test accuracy of pruned MPT-7B models on GPUs via N:M sparsity format.}
    \label{tab:NM_mpt}
    }
\end{minipage}
\hfill ~~~% to create some space between the two tables
\begin{minipage}{0.63\textwidth}
\centering
    \small{
    \setlength\tabcolsep{1.5pt}
    \begin{tabular}{l|c|c|ccc|ccc}
        \toprule
        Machine & \makecell{Num\\Cores} & \makecell{Dense\\FP32} & \makecell{60\%\\FP32} & \makecell{70\%\\FP32} & \makecell{80\%\\FP32} & \makecell{60\%\\INT8} & \makecell{70\%\\INT8} & \makecell{80\%\\INT8} \\
        \midrule
        \multirow{2}{*}{Xeon Gold 6430} & 1 & 0.6 & 1.2 & 1.5 & 2.1 & 3.9 & 4.7 & 6.1 \\
                                             & 4 & 2.1 & 4.3 & 5.5 & 8.2 & 14.1 & 16.3 & 19.6 \\
        \midrule
        \multirow{2}{*}{Ryzen 9 7950X} & 1 & 1.4 & 2.0 & 2.0 & 2.1 & 7.2 & 7.7 & 7.9 \\
                                           & 4 & 2.5 & 5.4 & 6.3 & 7.6 & 17.4 & 20.9 & 26.7 \\
        \bottomrule
    \end{tabular}
    \captionof{table}{Latency performance (tokens/second) for the MPT-7B model at various sparsities, using DeepSparse~\cite{deepsparse} on 1 and 4 cores, using Intel and AMD CPUs.}
    \label{tab:TPS}
    }
\end{minipage}

\paragraph{Discussion.} 
The accuracy results for different losses are shown in Figure~\ref{fig:mpt-comparison}. They exhibit very similar trends to the prior two applications, showing that SquareHead KD is superior to both standard CE loss and standard KD for sparse fine-tuning. The fact that the distilled models can outperform the dense baseline at low sparsity can be explained due to the effect of distillation being applied, but also possibly the regularizing effect of low sparsity. 
We observe that, with SquareHead KD, we can obtain FP32 models with 70\% and 75\% sparsities which have essentially no loss (in terms of test accuracy) relative to the dense model, even though pruning is performed in one shot. It is worth emphasizing that fine-tuning with SquareHead KD improves the dense model's accuracy as well, from $28.2$ to $33.0$.

We now examine the speedup-vs-accuracy trade-offs, presented in Table~\ref{tab:compression_mpt} for both FP32 and INT8 models. 
In FP32, moderate sparsities (60-70\%) can be reached losslessly, leading to speedups of 2-2.5x, whereas higher sparsities (80\%) lead to higher accuracy loss. 
Moving to INT8, we observe a consistent accuracy decrease of 1-2\% at each sparsity level, due to post-training quantization. 
At the same time, this loss of accuracy is accompanied by a major performance improvement, since the gains due to these two compression approaches compound. 
In absolute terms, the 70\% INT8 model can execute at  a remarkable \textbf{7.7 tokens/second on a single core} of an AMD Ryzen CPU, and at \textbf{20.9 tokens/second} on 4 cores. 
The speedup of the lossless 60\%-sparse INT8 model is of approximately 6.7x, and we can reach a decoding speedup of 9.08x at 80\% INT8, at the price of a 7\% drop in accuracy.

% \vspace{-1em}
\section{Discussion}
% \vspace{-0.5em}

We have shown early  results suggesting that sparsity can be an effective acceleration approach for LLM inference, focusing on ``modern'' applications such as speech and language translation, and text generation following complex instructions. 
Importantly, our study shows that sparsity is compatible with quantization in the memory-bound generative setting, and that together these techniques can lead to remarkable performance for computationally-limited devices. 
 Based on prior work, we have identified a general distillation approach to recover accuracy while inducing high sparsity over limited fine-tuning data. 
 In future work, we plan to expand this preliminary study by exploring sparse fine-tuning for larger-scale models and tasks in the generative setting, higher quantization degrees, as well as the practically-relevant setting of pruning on the pretraining task~\cite{frantar2023scaling}, followed by fine-tuning the already-sparsified model on a specialized dataset.  

\section{Reproducibility}
\label{sec:reproducibility}

To promote reproducibility of our results, we provide the following resources: 

\begin{itemize}
    \item Code for sparse fine-tuning of T5 and Whisper models can be found here:~\url{https://github.com/IST-DASLab/TACO4NLP}. 
    \item Code for sparse fine-tuning of GPT-type (e.g. MPT) models can be found here:~\url{https://github.com/IST-DASLab/SparseFinetuning}.
    \item MPT models are released at~\url{https://sparsezoo.neuralmagic.com/?datasets=gsm8k&ungrouped=true}
    \item CPU speedup for generative inference can be reproduced by following the instructions at \url{https://github.com/neuralmagic/deepsparse/tree/main/research/mpt}  and as a \href{https://huggingface.co/collections/neuralmagic/sparse-finetuning-mpt-65241d875b29204d6d42697d}{HuggingFace Collection}. 
    \item GPU code will be provided at \url{https://github.com/IST-DASLab/sparsegpt/}. 
\end{itemize}

\section*{Acknowledgments}

The authors thank the Neural Magic Machine Learning Research team for useful discussions during the development of this paper. We would also like to thank Eugenia Iofinova for useful comments on an earlier version of this draft, and Artur Niederfahrenhorst for useful suggestions regarding fine-tuning on the GSM dataset.  

%Bibliography
\bibliographystyle{unsrt}  
\bibliography{references}  

\begin{thebibliography}{10}

\bibitem{vaswani2017attention}
Ashish Vaswani, Noam Shazeer, Niki Parmar, Jakob Uszkoreit, Llion Jones, Aidan~N Gomez, {\L}ukasz Kaiser, and Illia Polosukhin.
\newblock Attention is all you need.
\newblock In {\em Conference on Neural Information Processing Systems (NeurIPS)}, 2017.

\bibitem{dao2022flashattention}
Tri Dao, Daniel~Y Fu, Stefano Ermon, Atri Rudra, and Christopher R{\'e}.
\newblock {FlashAttention}: Fast and memory-efficient exact attention with io-awareness.
\newblock {\em arXiv preprint arXiv:2205.14135}, 2022.

\bibitem{frantar2022gptq}
Elias Frantar, Saleh Ashkboos, Torsten Hoefler, and Dan Alistarh.
\newblock Gptq: Accurate post-training quantization for generative pre-trained transformers.
\newblock {\em arXiv preprint arXiv:2210.17323}, 2022.

\bibitem{dettmers2022optimizers}
Tim Dettmers, Mike Lewis, Sam Shleifer, and Luke Zettlemoyer.
\newblock 8-bit optimizers via block-wise quantization.
\newblock {\em 9th International Conference on Learning Representations, ICLR}, 2022.

\bibitem{dettmers2022case}
Tim Dettmers and Luke Zettlemoyer.
\newblock The case for 4-bit precision: k-bit inference scaling laws.
\newblock {\em arXiv preprint arXiv:2212.09720}, 2022.

\bibitem{dettmers2022llm}
Tim Dettmers, Mike Lewis, Younes Belkada, and Luke Zettlemoyer.
\newblock {LLM}.int8(): 8-bit matrix multiplication for transformers at scale.
\newblock {\em Advances in Neural Information Processing Systems 35: Annual Conference on Neural Information Processing Systems 2022, NeurIPS 2022}, 2022.

\bibitem{xiao2022smoothquant}
Guangxuan Xiao, Ji~Lin, Mickael Seznec, Julien Demouth, and Song Han.
\newblock Smoothquant: Accurate and efficient post-training quantization for large language models.
\newblock {\em arXiv preprint arXiv:2211.10438}, 2022.

\bibitem{yao2022zeroquant}
Zhewei Yao, Reza~Yazdani Aminabadi, Minjia Zhang, Xiaoxia Wu, Conglong Li, and Yuxiong He.
\newblock Zeroquant: Efficient and affordable post-training quantization for large-scale transformers.
\newblock {\em arXiv preprint arXiv:2206.01861}, 2022.

\bibitem{dettmers2023spqr}
Tim Dettmers, Ruslan Svirschevski, Vage Egiazarian, Denis Kuznedelev, Elias Frantar, Saleh Ashkboos, Alexander Borzunov, Torsten Hoefler, and Dan Alistarh.
\newblock Spqr: A sparse-quantized representation for near-lossless llm weight compression.
\newblock {\em arXiv preprint arXiv:2306.03078}, 2023.

\bibitem{chee2023quip}
Jerry Chee, Yaohui Cai, Volodymyr Kuleshov, and Christopher De~Sa.
\newblock Quip: 2-bit quantization of large language models with guarantees.
\newblock {\em arXiv preprint arXiv:2307.13304}, 2023.

\bibitem{lecun1990optimal}
Yann LeCun, John~S Denker, and Sara~A Solla.
\newblock Optimal brain damage.
\newblock In {\em Conference on Neural Information Processing Systems (NeurIPS)}, 1990.

\bibitem{devlin2018bert}
Jacob Devlin, Ming-Wei Chang, Kenton Lee, and Kristina Toutanova.
\newblock {BERT}: Pre-training of deep bidirectional transformers for language understanding.
\newblock In {\em North American Chapter of the Association for Computational Linguistics (NAACL)}, 2019.

\bibitem{2020-sanh}
Victor Sanh, Thomas Wolf, and Alexander~M. Rush.
\newblock Movement pruning: Adaptive sparsity by fine-tuning.
\newblock {\em arXiv preprint arXiv:2005.07683}, 2020.

\bibitem{kurtic2022optimal}
Eldar Kurtic, Daniel Campos, Tuan Nguyen, Elias Frantar, Mark Kurtz, Benjamin Fineran, Michael Goin, and Dan Alistarh.
\newblock The {Optimal BERT Surgeon}: Scalable and accurate second-order pruning for large language models.
\newblock {\em arXiv preprint arXiv:2203.07259}, 2022.

\bibitem{radford2023robust}
Alec Radford, Jong~Wook Kim, Tao Xu, Greg Brockman, Christine McLeavey, and Ilya Sutskever.
\newblock Robust speech recognition via large-scale weak supervision.
\newblock In {\em International Conference on Machine Learning}, pages 28492--28518. PMLR, 2023.

\bibitem{wei2021finetuned}
Jason Wei, Maarten Bosma, Vincent~Y Zhao, Kelvin Guu, Adams~Wei Yu, Brian Lester, Nan Du, Andrew~M Dai, and Quoc~V Le.
\newblock Finetuned language models are zero-shot learners.
\newblock {\em arXiv preprint arXiv:2109.01652}, 2021.

\bibitem{machavcek2014results}
Matou{\v{s}} Mach{\'a}{\v{c}}ek and Ond{\v{r}}ej Bojar.
\newblock Results of the wmt14 metrics shared task.
\newblock In {\em Proceedings of the Ninth Workshop on Statistical Machine Translation}, pages 293--301, 2014.

\bibitem{mpt}
MosaicML~NLP Team.
\newblock Introducing mpt-7b: A new standard for open-source, commercially usable llms.
\newblock {\em www.mosaicml.com/blog/mpt-7b}, 2023.
\newblock Accessed: 2023-10-01.

\bibitem{cobbe2021training}
Karl Cobbe, Vineet Kosaraju, Mohammad Bavarian, Mark Chen, Heewoo Jun, Lukasz Kaiser, Matthias Plappert, Jerry Tworek, Jacob Hilton, Reiichiro Nakano, et~al.
\newblock Training verifiers to solve math word problems.
\newblock {\em arXiv preprint arXiv:2110.14168}, 2021.

\bibitem{frantar2023massive}
Elias Frantar and Dan Alistarh.
\newblock Massive language models can be accurately pruned in one-shot.
\newblock {\em arXiv preprint arXiv:2301.00774}, 2023.

\bibitem{hinton2015distilling}
Geoffrey Hinton, Oriol Vinyals, and Jeff Dean.
\newblock Distilling the knowledge in a neural network.
\newblock {\em arXiv preprint arXiv:1503.02531}, 2015.

\bibitem{sun2019patient}
Siqi Sun, Yu~Cheng, Zhe Gan, and Jingjing Liu.
\newblock Patient knowledge distillation for bert model compression.
\newblock {\em arXiv preprint arXiv:1908.09355}, 2019.

\bibitem{kurtic2023ziplm}
Eldar Kurtic, Elias Frantar, and Dan Alistarh.
\newblock Ziplm: Hardware-aware structured pruning of language models.
\newblock {\em arXiv preprint arXiv:2302.04089}, 2023.

\bibitem{frantar2022spdy}
Elias Frantar and Dan Alistarh.
\newblock {SPDY:} {A}ccurate pruning with speedup guarantees.
\newblock {\em arXiv preprint arXiv:2201.13096}, 2022.

\bibitem{deepsparse}
NeuralMagic.
\newblock {DeepSparse}.
\newblock https://github.com/neuralmagic/deepsparse, 2021.

\bibitem{xia2022structured}
Mengzhou Xia, Zexuan Zhong, and Danqi Chen.
\newblock Structured pruning learns compact and accurate models.
\newblock {\em arXiv preprint arXiv:2204.00408}, 2022.

\bibitem{gale2020sparse}
Trevor Gale, Matei Zaharia, Cliff Young, and Erich Elsen.
\newblock Sparse gpu kernels for deep learning.
\newblock In {\em SC20: International Conference for High Performance Computing, Networking, Storage and Analysis}, pages 1--14. IEEE, 2020.

\bibitem{hoefler2021sparsity}
Torsten Hoefler, Dan Alistarh, Tal Ben-Nun, Nikoli Dryden, and Alexandra Peste.
\newblock Sparsity in deep learning: Pruning and growth for efficient inference and training in neural networks.
\newblock {\em arXiv preprint arXiv:2102.00554}, 2021.

\bibitem{raffel2020exploring}
Colin Raffel, Noam Shazeer, Adam Roberts, Katherine Lee, Sharan Narang, Michael Matena, Yanqi Zhou, Wei Li, and Peter~J. Liu.
\newblock Exploring the limits of transfer learning with a unified text-to-text transformer, 2020.

\bibitem{bojar-etal-2014-findings}
Ond{\v{r}}ej Bojar, Christian Buck, Christian Federmann, Barry Haddow, Philipp Koehn, Johannes Leveling, Christof Monz, Pavel Pecina, Matt Post, Herve Saint-Amand, Radu Soricut, Lucia Specia, and Ale{\v{s}} Tamchyna.
\newblock Findings of the 2014 workshop on statistical machine translation.
\newblock In {\em Proceedings of the Ninth Workshop on Statistical Machine Translation}, pages 12--58, Baltimore, Maryland, USA, June 2014. Association for Computational Linguistics.

\bibitem{radford2022robust}
Alec Radford, Jong~Wook Kim, Tao Xu, Greg Brockman, Christine McLeavey, and Ilya Sutskever.
\newblock Robust speech recognition via large-scale weak supervision, 2022.

\bibitem{ardila-etal-2020-common}
Rosana Ardila, Megan Branson, Kelly Davis, Michael Kohler, Josh Meyer, Michael Henretty, Reuben Morais, Lindsay Saunders, Francis Tyers, and Gregor Weber.
\newblock Common voice: A massively-multilingual speech corpus.
\newblock In {\em Proceedings of the Twelfth Language Resources and Evaluation Conference}, pages 4218--4222, Marseille, France, May 2020. European Language Resources Association.

\bibitem{gsm8k}
Karl Cobbe, Vineet Kosaraju, Mohammad Bavarian, Mark Chen, Heewoo Jun, Lukasz Kaiser, Matthias Plappert, Jerry Tworek, Jacob Hilton, Reiichiro Nakano, et~al.
\newblock Training verifiers to solve math word problems.
\newblock {\em arXiv preprint arXiv:2110.14168}, 2021.

\bibitem{Anyscale}
Artur Niederfahrenhorst, Kourosh Hakhamaneshi, and Rehaan Ahmad.
\newblock {Fine-Tuning LLMs: LoRA or Full-Parameter?}, 2021.

\bibitem{pmlr-v119-kurtz20a}
Mark Kurtz, Justin Kopinsky, Rati Gelashvili, Alexander Matveev, John Carr, Michael Goin, William Leiserson, Sage Moore, Bill Nell, Nir Shavit, and Dan Alistarh.
\newblock Inducing and exploiting activation sparsity for fast inference on deep neural networks.
\newblock In {\em International Conference on Machine Learning (ICML)}, 2020.

\bibitem{lmeval}
Leo Gao, Jonathan Tow, Stella Biderman, Sid Black, Anthony DiPofi, Charles Foster, Laurence Golding, Jeffrey Hsu, Kyle McDonell, Niklas Muennighoff, Jason Phang, Laria Reynolds, Eric Tang, Anish Thite, Ben Wang, Kevin Wang, and Andy Zou.
\newblock A framework for few-shot language model evaluation.
\newblock {\em https://github.com/EleutherAI/lm-evaluation-harness}, 2021.

\bibitem{frantar2023scaling}
Elias Frantar, Carlos Riquelme, Neil Houlsby, Dan Alistarh, and Utku Evci.
\newblock Scaling laws for sparsely-connected foundation models.
\newblock {\em arXiv preprint arXiv:2309.08520}, 2023.

\end{thebibliography}

\,
\appendix
\,

\newpage
\section{Sparse Fine-tuning Hyperparameters}
\label{appendix:training_hyperparameters}

\subsection{Parameters for Whisper and T5 Experiments}
In Table~\ref{tab:training_hyperparameters} we report hyperparameters for sparse fine-tuning of Whisper and T5 models.

\begin{table}[!h]
  \centering
   \caption{Hyperparameters used in the sparse fine-tuning experiments with Whisper and T5 models.}
   \small{
       \setlength\tabcolsep{1.5pt}
      \begin{tabular}{ccccccc}
       \toprule
       Model & Learning rate & Num epochs & Warmup steps & Batch size & Schedule & Weight decay \\
       \midrule
       T5-Small & $2 \cdot 10^{-3}$ & 3 & 300 & 128 & linear & $10^{-4}$\\
       \midrule
       Whisper-Small & $2 \cdot 10^{-4}$ & 6 & 50 & 32 & linear & $10^{-4}$ \\
      \bottomrule
      \end{tabular}
  }
  \label{tab:training_hyperparameters}
\end{table}

\subsection{Parameters for MPT Experiments}
\label{appendix:mpt_hyperparameters}
For MPT-7B model, we untie input embeddings and language modelling head for compatibility with quantization via SparseML, where we quantize weights of input embeddings, and weights and activations of the language modelling head. For all sparsities, we oneshot prune the model with the default SparseGPT~\cite{frantar2023massive} parameters and then fine-tune for either 2 (40\%, 50\%, and 60\% sparsity) or 4 epochs (70\% and 80\% sparsity). We use linearly decaying learning rate (LR) with warmup of 20 steps, and sweep for all loss variants over the following peak-LR values: 3e-5, 5e-5, 8e-5, 1e-4. We use batch-size of 32, and equal weight $\lambda = 1.0$ for losses in compound loss variants. For more details on reproducibility, please check our GitHub repository ~\url{https://github.com/IST-DASLab/SparseFinetuning}.

\section{Entropy Analysis}
\label{appendix:entropy}
A possible explanation for the success of SquareHead could be the fact, that fine-tuning with task loss only, especially on limited data, leads to overfitting. We measured the entropy of the predictive distribution
of sparse Whisper models and observed that models fine-tuned without knowledge distillation have very low entropy, i.e make overconfident predictions, whereas SquareHead induces regularization. 

\begin{figure}[!h]
    \centering
    \includegraphics[width=0.8\linewidth]{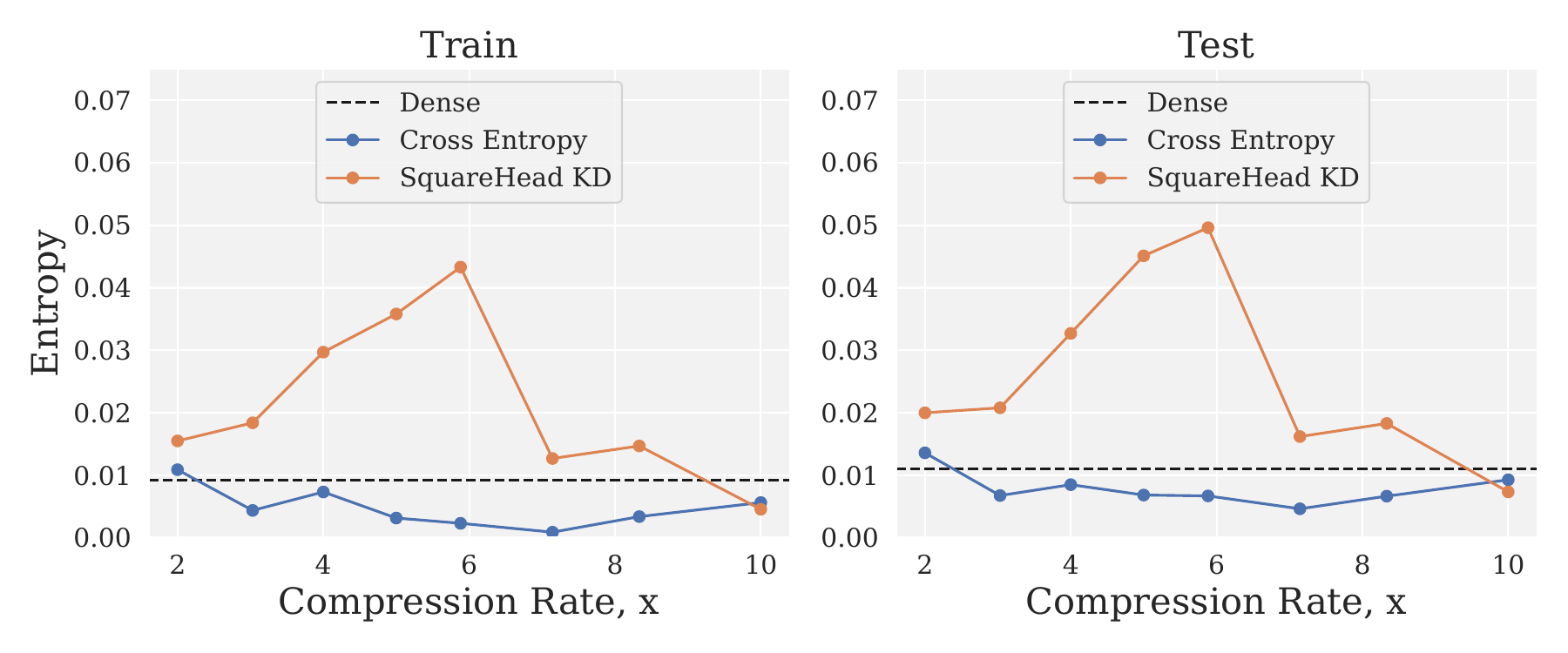}
    \captionof{figure}{Entropy of the predictive distribution of sparse Whisper-244M models for different fine-tuning approaches 
    on train and test split of Hindi subset of Common Voice dataset.}
    \label{fig:asr_entropy_comparison}
\end{figure}

\section{Divergence in training T5 on WMT14 dataset}\label{appendix:t5_divergence}

\begin{figure}[!h]
    \centering
    \includegraphics[width=0.4\linewidth]{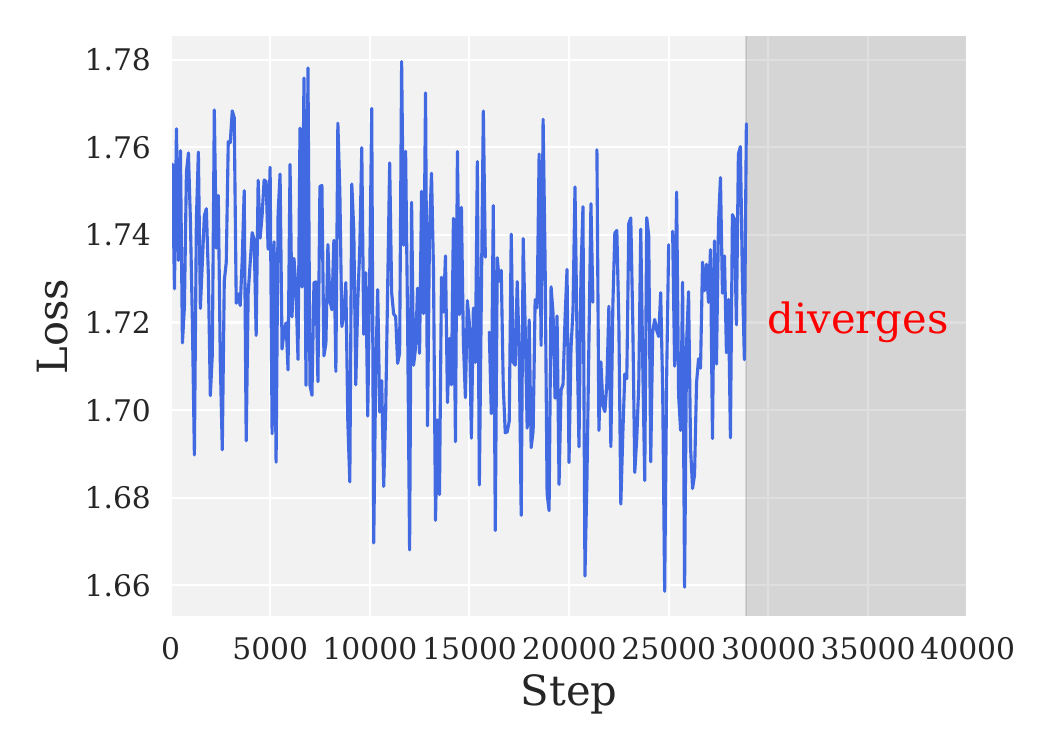}
    \captionof{figure}{Loss history for diverged T5 run.}
    \label{fig:t5_divergence_example}
\end{figure}

We observed that sparse fine-tuning of T5 becomes unstable at high sparsity when training with task loss only or standard KD. 
Training loss after pruning step decreases for some number of steps and then suddenly diverges as shown on Figure \ref{fig:t5_divergence_example}. This issue is encountered even when trying different random seeds. Behavior is similar for training with CE and standard KD. Only training with SquareHead loss appears to be stable at high sparsity.  

\section{T5 and Whisper Encoder-Decoder Performance Breakdown}
\label{appendix:t5_whisper_perf}
\begin{figure}[!h]
    \centering
    
    % First minipage for the figure
    \begin{minipage}[b]{0.32\linewidth}
        \centering
        \includegraphics[width=0.9\linewidth]{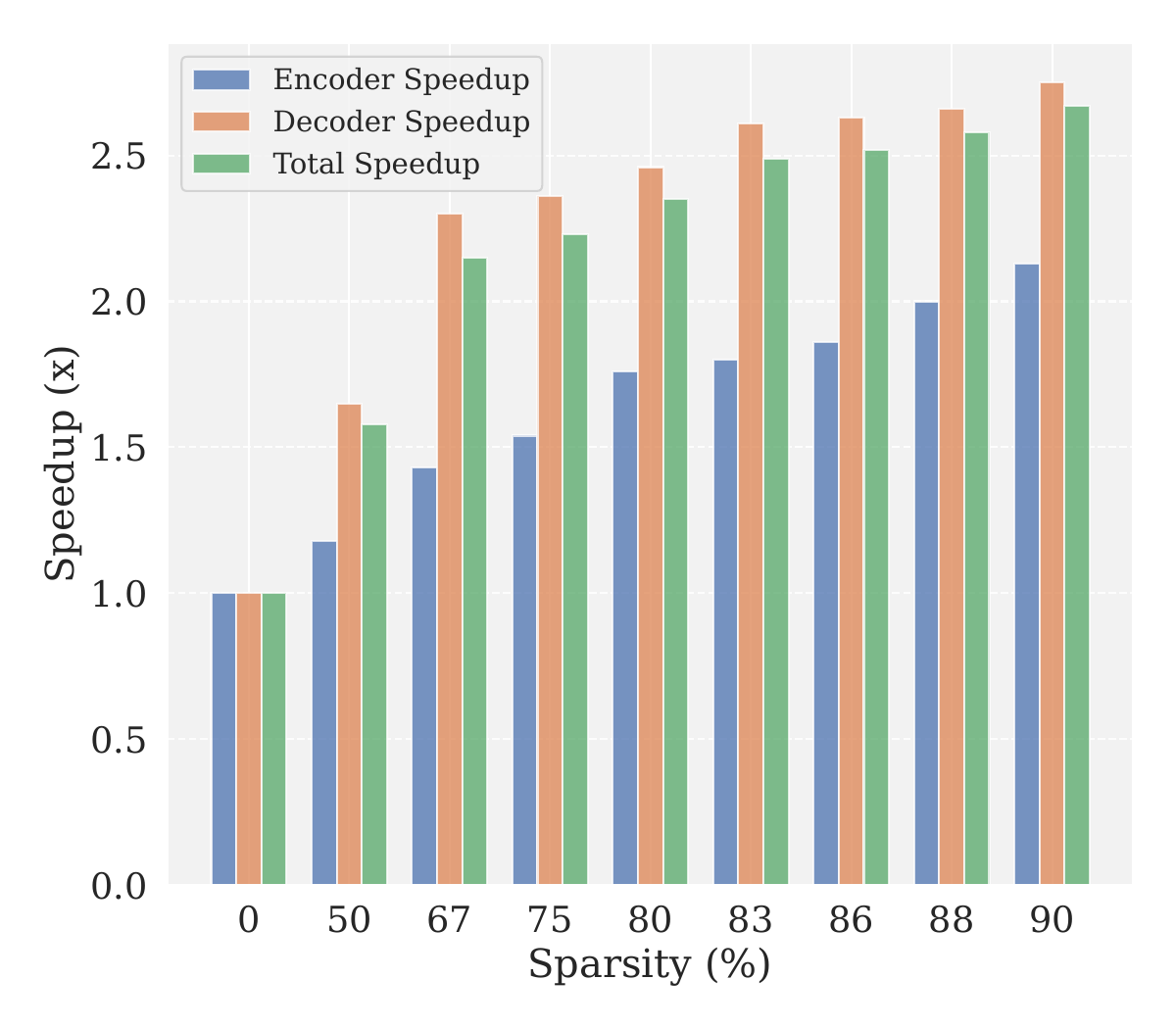}
        \captionof{figure}{Component speedups of Whisper-Small for various sparsities.}
        \label{fig:whisper_component_speedup}
    \end{minipage}
    \hfill 
    % Creates a horizontal space between the minipages
    % Second minipage for the table
    \begin{minipage}[b]{0.65\linewidth}
        \centering
        \small{
        \setlength\tabcolsep{1.7pt}
        \begin{tabular}{c|cc|cc|cc}
            \toprule & \multicolumn{2}{c|}{Encoder} & \multicolumn{2}{c}{Decoder} & \multicolumn{2}{|c}{Total} \\
            \midrule
            \makecell{Sparsity\\(\%)} & \makecell{Speedup} &  \makecell{Latency} & \makecell{Speedup} &  \makecell{Latency} & \makecell{Speedup} &  \makecell{Latency} \\
            \midrule
            0  & 1.00x & 89.39 ms & 1.00x & 12.82 ms & 1.00x & 882.70 ms \\
            50 & 1.18x & 75.49 ms & 1.65x &  7.76 ms & 1.58x & 558.21 ms \\
            67 & 1.43x & 62.50 ms & 2.30x &  5.57 ms & 2.15x & 410.06 ms \\
            75 & 1.54x & 58.03 ms & 2.36x &  5.43 ms & 2.23x & 395.53 ms \\
            80 & 1.76x & 50.87 ms & 2.46x &  5.21 ms & 2.35x & 374.93 ms \\
            83 & 1.80x & 49.77 ms & 2.61x &  4.91 ms & 2.49x & 354.59 ms \\
            86 & 1.86x & 48.01 ms & 2.63x &  4.87 ms & 2.52x & 350.16 ms \\
            88 & 2.00x & 44.66 ms & 2.66x &  4.81 ms & 2.58x & 342.70 ms \\
            90 & 2.13x & 41.97 ms & 2.75x &  4.66 ms & 2.67x & 330.35 ms \\
            \bottomrule
        \end{tabular}
        }
        \captionof{table}{Encoder, decoder, and total timings of Whisper-Small for various sparsities. The total workload is generating 60 tokens from 15 seconds of audio.} 
        \label{tab:whisper_breakdown}
    \end{minipage}

\end{figure}

\begin{figure}[!h]
    \centering
    
    % First minipage for the figure
    \begin{minipage}[b]{0.32\linewidth}
        \centering
        \includegraphics[width=0.9\linewidth]{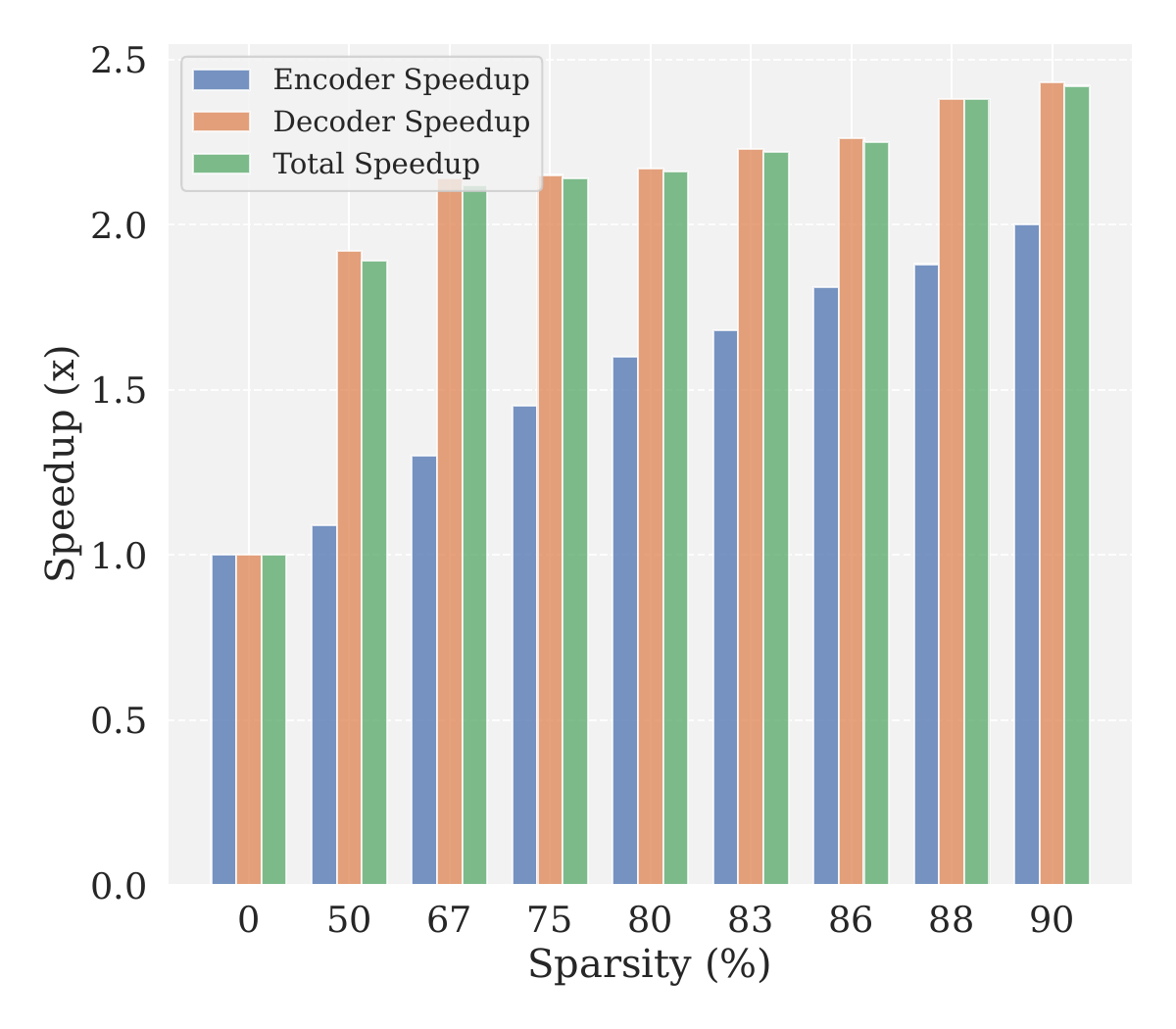}
        \captionof{figure}{Component speedups of T5-Small for various sparsities.}
        \label{fig:t5_component_speedup}
    \end{minipage}
    \hfill 
    % Creates a horizontal space between the minipages
    % Second minipage for the table
    \begin{minipage}[b]{0.65\linewidth}
        \centering
        \small{
        \setlength\tabcolsep{1.7pt}
        \begin{tabular}{c|cc|cc|cc}
            \toprule & \multicolumn{2}{c|}{Encoder} & \multicolumn{2}{c}{Decoder} & \multicolumn{2}{|c}{Total} \\
            \midrule
            \makecell{Sparsity\\(\%)} & \makecell{Speedup} &  \makecell{Latency} & \makecell{Speedup} &  \makecell{Latency} & \makecell{Speedup} &  \makecell{Latency} \\
            \midrule
            0  & 1.00x & 5.18 ms & 1.00x & 2.99 ms & 1.00x & 391.92 ms \\
            50 & 1.09x & 4.75 ms & 1.92x & 1.56 ms & 1.89x & 206.93 ms \\
            67 & 1.30x & 3.97 ms & 2.14x & 1.40 ms & 2.12x & 184.80 ms \\
            75 & 1.45x & 3.58 ms & 2.15x & 1.39 ms & 2.14x & 183.51 ms \\
            80 & 1.60x & 3.23 ms & 2.17x & 1.38 ms & 2.16x & 181.18 ms \\
            83 & 1.68x & 3.08 ms & 2.23x & 1.34 ms & 2.22x & 176.32 ms \\
            86 & 1.81x & 2.86 ms & 2.26x & 1.33 ms & 2.25x & 174.21 ms \\
            88 & 1.88x & 2.75 ms & 2.38x & 1.26 ms & 2.38x & 164.96 ms \\
            90 & 2.00x & 2.59 ms & 2.43x & 1.23 ms & 2.42x & 161.84 ms \\
            \bottomrule
        \end{tabular}
        }
        \captionof{table}{Encoder, decoder, and total timings of T5-Small for various sparsities. The total workload is encoding 128 tokens and generating 128 tokens.} 
        \label{tab:t5_breakdown}
    \end{minipage}

\end{figure}

Understanding the individual performance of the encoder and decoder components in sequence-to-sequence models like T5 and Whisper is paramount. These components handle different tasks; the encoder processes input data into a context-rich representation, and the decoder translates this representation into generated tokens, using both the encoder's representation and it's own context to produce subsequent output. While individual component performance provides insight into bottlenecks, it's also essential to evaluate end-to-end performance as it offers a more holistic understanding of how compression affects real-world application scenarios. In Figures~\ref{fig:whisper_component_speedup} and ~\ref{fig:t5_component_speedup}, and Tables~\ref{tab:whisper_breakdown} and~\ref{tab:t5_breakdown} we present detailed performance breakdown from the DeepSparse engine.

\section{T5 inference examples}
\label{appendix:t5_and_whisper_inference}

Below in Table~\ref{tab:t5_inference_example} we present a couple of non cherry-picked examples of
English-German translation with dense and sparse T5 models.
Sparse model is pruned to 75\%. One can see, that sparse models yields outputs of superficially the same quality at the base dense model. 

\begin{table}[!h]
  \centering
   \caption{
   Examples of T5 English-German translation.}
   \small{
   \begin{tabular}{p{4cm}|p{4cm}|p{4cm}}
    \toprule
    Prompt & Dense model & Sparse model \\
    \midrule
     The quick brown fox jumps over the lazy dog. & Der braune Fuchs springt über den faulen Hund. & Der schnelle braune Fuchs springt über den faulen Hund. \\
    \midrule
     I don't respect anybody who can't tell the difference between Pepsi and Coke. & Ich respektiere niemanden, der den Unterschied zwischen Pepsi und Coke nicht erzählen kann. & Ich respektiere niemanden, der den Unterschied zwischen Pepsi und Coke nicht erzählen kann. \\
     \midrule
     Sometimes it is better to just walk away from things and go back to them later when you're in a better frame of mind. & Manchmal ist es besser, einfach weg von den Dingen zu gehen und zu ihnen später zurück, wenn Sie in einem besseren Geistesrahmen sind. & Manchmal ist es besser, sich von den Dingen zu entfernen und später zu ihnen zurückzukehren, wenn Sie in einem besseren Geistesrahmen sind. \\
    \bottomrule
    \end{tabular}
  }
  \label{tab:t5_inference_example}
\end{table}

\end{document}